\documentclass[conference]{IEEEtran}

\usepackage{times}  
\usepackage{helvet} 
\usepackage{courier}  
\usepackage[hyphens]{url}  
\usepackage{graphicx} 
\usepackage{amsmath}
\usepackage{array}
\usepackage{footnote}
\usepackage{tabularx}
\usepackage{multirow}
\usepackage{graphicx}  
\ifCLASSINFOpdf
\else
\fi
\begin{document}
%
\title{Few-Features Attack to Fool Machine Learning Models through Mask-Based GAN}

\author{\IEEEauthorblockN{Feng~CHEN}
\IEEEauthorblockA{Institute of Automation\\
Chinese Academy of Sciences, China\\
Email: chenfeng@ia.ac.cn}
\and
\IEEEauthorblockN{Yunkai~SHANG}
\IEEEauthorblockA{College of Electrical and Information Engineering\\
Hunan University, China\\
Email: sykhnu@126.com}
\and
\IEEEauthorblockN{Bo~XU}
\IEEEauthorblockA{Institute of Automation\\
Chinese Academy of Sciences, China\\
Email: boxu@ia.ac.cn}
\and
\IEEEauthorblockN{Jincheng~HU}
\IEEEauthorblockA{Institute of Automation\\
Chinese Academy of Sciences, China\\
Email: jincheng.hu@ia.ac.cn}
}


%


\maketitle

\begin{abstract}
GAN (Generative Adversarial Networks, \cite{goodfellow2014generative}) is a deep-learning based generative approach to generate contents such as images, languages and speeches. Recently, studies have shown that GAN can also be applied to generative adversarial attack examples (\cite{xiao2018generating}, \cite{lin2018idsgan}) to fool the machine-learning models.  In comparison with the previous non-learning adversarial example attack approaches, the GAN-based adversarial attack example approach can generate the adversarial samples quickly using the GAN architecture every time facing a new sample after training, but meanwhile needs to perturb the attack samples in great quantities, which results in the unpractical application in reality. To address this issue, we propose a new approach, named Few-Feature-Attack-GAN (FFA-GAN). FFA-GAN has a significant time-consuming advantage than the non-learning adversarial samples approaches and a better non-zero-features performance than the GAN-based adversarial sample approaches. FFA-GAN can automatically generate the attack samples in the black-box attack through the GAN architecture instead of the evolutional algorithms or the other non-learning approaches. Besides, we introduce the mask mechanism into the generator network of the GAN architecture to optimize the constraint issue, which can also be regarded as the sparsity problem of the important features. During the training, the different weights of losses of the generator are set in the different training phases to ensure the divergence of the two above mentioned parallel networks of the generator. Experiments are made respectively on the structured data sets KDD-Cup 1999 and CIC-IDS 2017, in which the dimensions of the data are relatively low, and also on the unstructured data sets MNIST and CIFAR-10 with the data of the relatively high dimensions. The results of the experiments demonstrate the effectiveness and the robustness of our proposed approach.
\end{abstract}


%

\section{I. Introduction}
\noindent GAN (Generative Adversarial Networks, \cite{goodfellow2014generative}) is an unsupervised learning method and a deep model architecture which consists of a generative model and a discriminative model. The principle of GAN is to build a zero-sum game through the competition of two neural networks and then two networks compete each other to reach Nash equilibrium to learn to create the complex distribution of the instance. Recently, GAN is more and more widely used in many domains, not only limitedly in the fields of the generation of images (\cite{goodfellow2014generative}, \cite{arjovsky2017wasserstein}, \cite{mao2017least}), speeches \cite{pascual2017segan} and languages \cite{yu2017seqgan} , but also in the other fields like adversarial examples (\cite{xiao2018generating}, \cite{lin2018idsgan}) and malware generation \cite{hu2017generating}. This paper is going to discuss the application of an improved GAN method to generate attack adversarial examples to attack the machine-learning based models, especially concerning about the issue of reducing the perturbation when producing the attack adversarial examples using the GAN architecture.

Adversarial example is a sample example which has been modified additionally, and its goal is to make the machine learning models to misclassify it. There are previously many studies in the field of generative attack example. \cite{szegedy2013intriguing} explained the concept of adversarial examples. It refers to an input sample formed by deliberately adding subtle perturbation in the data set, which causes the model to give a false prediction with high confidence. \cite{goodfellow2014explaining} proposed fast gradient sign method (FGSM) that can produce the subtle perturbation, and it is a white-box attack method. JSMA\cite{papernot2016limitations}is also a white-box attack method, which calculates $L_2$ and $L_{\infty}$ to build the saliency map to limit the perturbation in a small range. \cite{su2019one} proposed a black-box attack approach which can use only very few pixels to implement the adversarial sample attack using the evolutional algorithm. The drawback of the optimization-based approaches is that these approaches are not the learning methods so that we have to consume too much time to calculate the perturbation for the attack sample when facing a new attack sample.

For the learning approach, GAN is a well-behaved alternative in the adversarial sample because of its good generative ability. \cite{xiao2018generating} proposed a method (AdvGAN) to generate adversarial examples through GAN. They implemented both of the white-box and the black-box attack method, the method only was used in image processing. \cite{lin2018idsgan} proposed IDSGAN that can generate adversarial attacks and they generated adversarial malicious traffic for the IDS (Instrusion Detection System) data set through a black attack approach, but they completely replaced almost all features. These two approaches both require lots of features being changed when attacking.
\begin{figure}[h]
    \includegraphics[width=80mm]{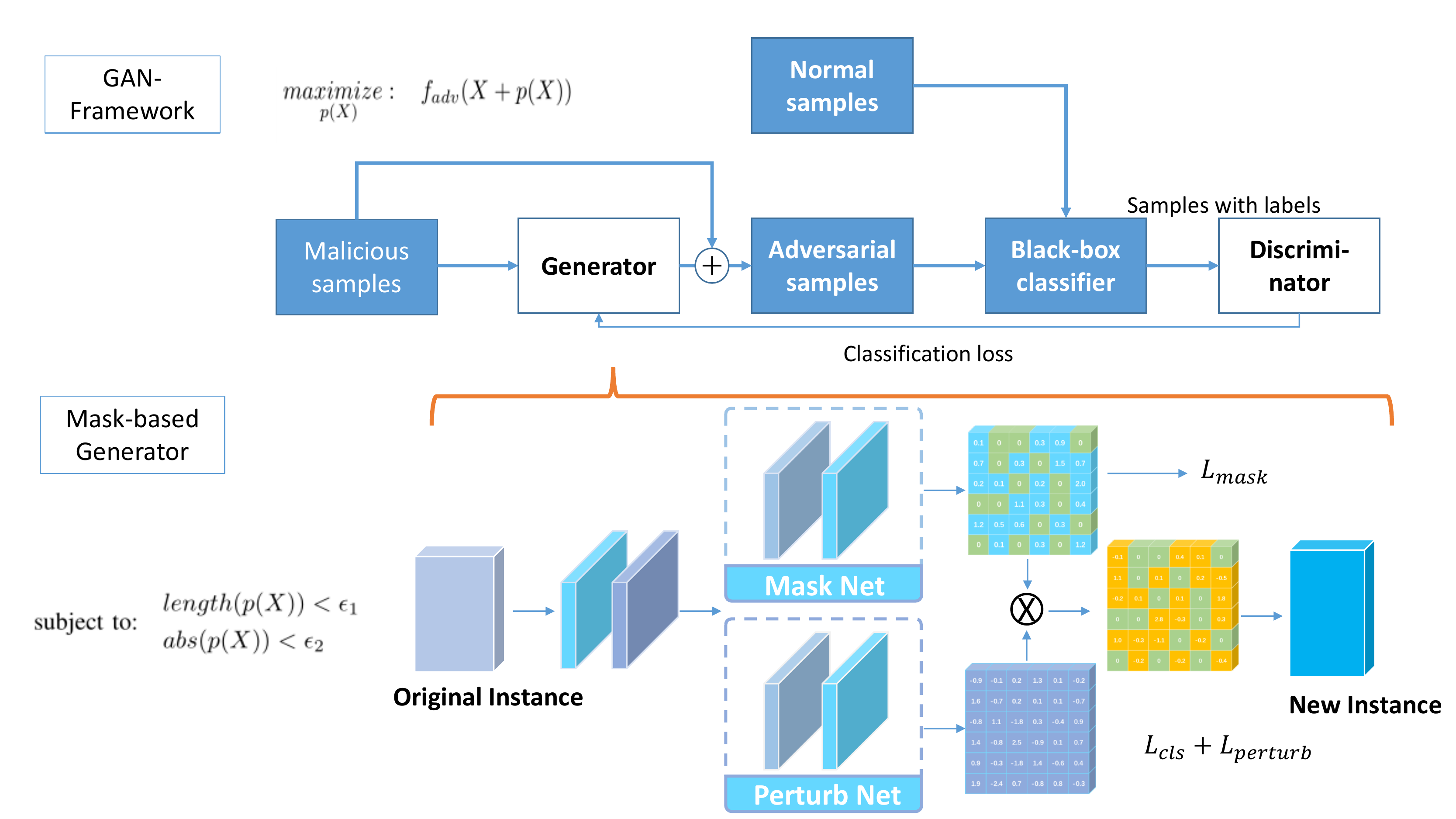}
    \label{FFA-GAN}
    \caption{FFA-GAN}
\end{figure}

In this paper, we propose a new improved GAN architecture named Few-Feature-Attack-GAN (FFA-GAN) to implement the adversarial attack to produce perturbations which are added in the attack instance to fool machine learning classifier models. Instead of using the great mass of features of the original attack instance, only few features of the original attack instance need to be changed because of the introduction of the mask mechanism in the generator network. The generator is composed of two parallel networks, the mask network and the perturbation network. The mask network defines which features of the changeable features on the instance can be perturbed during the training, and the perturbation network provides the perturbation of the all changeable features on the instance. Through the matrix dot production of two outputs from the mask network and the perturbation network, we can get the generator output as the perturbation (padding is needed if the dimensions of generator output are smaller than the all dimensions of the sample). The losses of the generator are composed of different parts, including the classification loss, the perturbation loss and the mask loss and the loss of the discriminator is the classification loss for the instance. With the help of the GAN framework, the generator and the discriminator build a zero-sum game to compete with each other, and reach an equilibrium to get the final perturbation with the few changed featured and the high bypass rate at the end.
\begin{figure}[h]
    \centering
    \includegraphics[width=90mm]{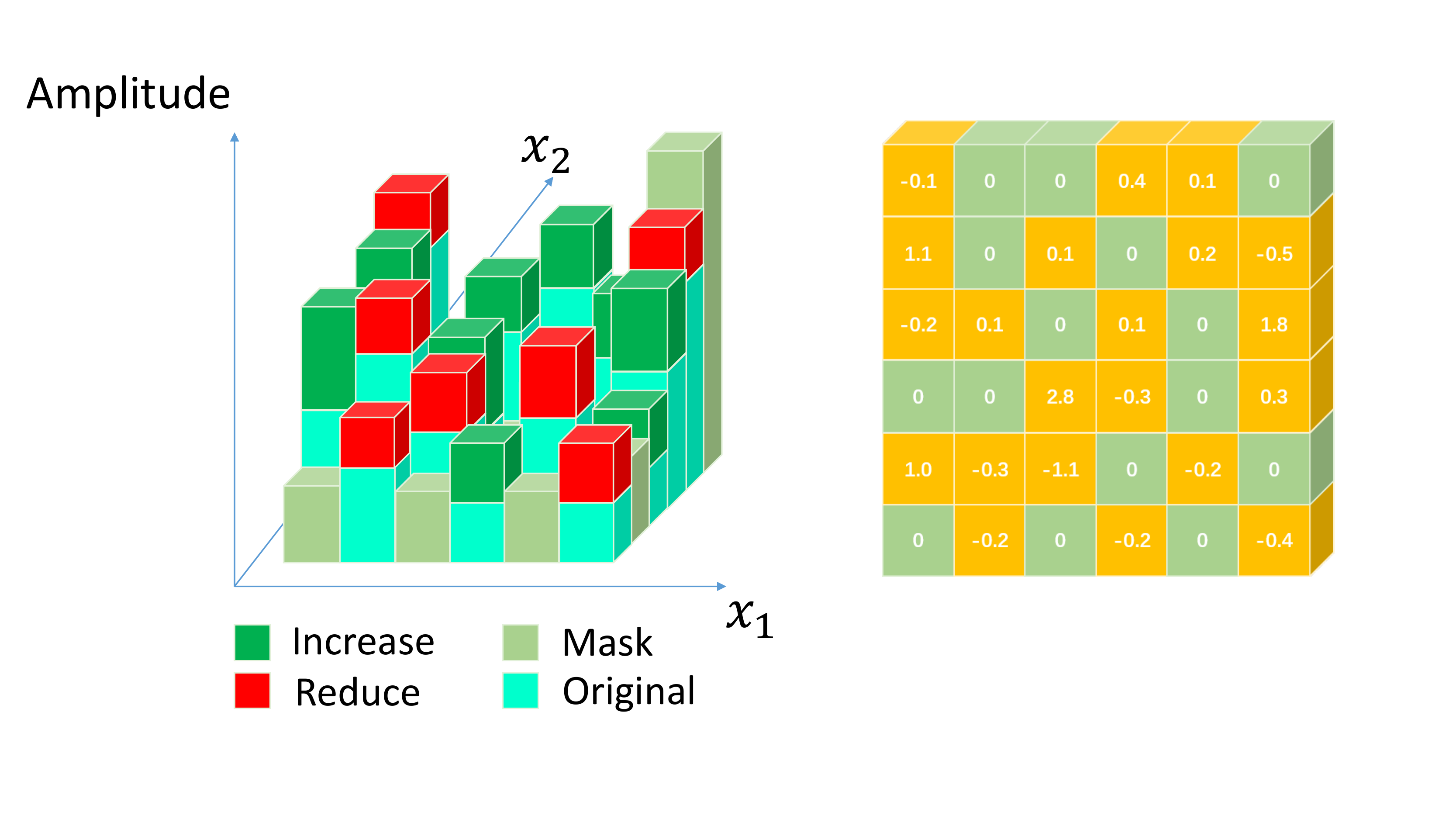}
    \label{Visualization of the Perturbation}
    \caption{Visualization of the Perturbation}
\end{figure}

To train the parallel networks, the several weight pairs for the losses should be adaptively changed so that the training can converge. At the beginning of the training, we set the weight of the classification loss large and the weight of the mask loss small. As the training goes, the weight of the classification decreases gradually and the weight of the mask loss decreases. We can regard the two phases as "the exploration phase" and "the exploitation phase" respectively. That means we first explore a good result in a relatively wide range and then optimize the range as small as possible to exploit a better result.

Our main contributions of the proposed FFA-GAN approach can be concluded as follows:
\\(1)In FFA-GAN, we introduce the mask mechanism into the generator network. The mask mechanism provides a filter to reduce the number of changed features or the perturbation of the samples and meanwhile without reducing the bypass rate through the GAN architecture.
\\(2)We define the losses of the generator in GAN which are composed of three parts: the classification loss, the loss of magnitude of perturbations, and the loss of non-zero-features' length of perturbations in the mask network.
\\(3)During the training, we use a specific training strategy to change the weights of different losses adaptively, and ensure the convergence of FFA-GAN and achieve a better performance than the method with the fixed loss weights.

\section{II. Preliminary}
\noindent We define generating adversarial attack samples to a target machine-learning model as a constraint optimization problem. The task of an adversarial attack sample is to modify the original attack sample as little as possible to make the target model misclassify it as the normal one. Based on these, we assume a target machine-learning classifier model $L$, the sample can be expressed as a vector of scalar features $ X = \{ x_1, x_2,... \}$. The perturbation of the adversarial sample can be represented as $p(X)= \{ x_1',x_2',... \} $, which will be added in the original sample as the new sample.
 \begin{equation}
\begin{aligned}
 X'=X+p(X) =  \{ x_1 + x_1', x_2 +x_2',... \}
\end{aligned}
\end{equation}

The model $L$ can classify the sample if it is malicious or not or whether it is the target class. We use $f(X)$ to express the ability of the model to classify the attack sample as the attack sample. To describe the ability of the adversarial ability, we here define the goal function as the bypass rate as follows:
\begin{equation}
\begin{aligned}
 rate_{bypass} = max(1-f(X')/f(X),0)\\
\end{aligned}
\end{equation}

The goal of generating adversarial attack is to maximize the high bypass rate:
\begin{equation}
\begin{aligned}
& \mathop{maximize:}\limits_{p(X)} rate_{bypass}\\
\end{aligned}
\end{equation}

While maintaining the high bypass rate, the difference of the adversarial attack sample and the original attack sample should be as little as possible. Therefore the constraints of the task can be describe as the following form:
\begin{equation}
\text{subject to:}
\begin{aligned}
& \quad length(p(X))<\epsilon_{1}\\
& \quad abs(p(X))<\epsilon_{2}\\
\end{aligned}
\end{equation}

In the domain of the adversarial examples, untargeted attack is an attack that fools the model to predict the sample as the type which is not the designated. Here y is the attack class label.
\begin{equation}
\begin{aligned}
f(X') \ne y\\
\end{aligned}
\end{equation}

Correspondingly, the goal of the targeted attack is to predict the sample as the target type. Here t is the target class label.
\begin{equation}
\begin{aligned}
f(X')=t\\
\end{aligned}
\end{equation}

Black-box attack is an attack that the attacker does not know the details of the inner information of the target model, such as the parameters, the structures and so on. On the contrary, the every details of the target model are known to the attacker in the white-box attack. We discuss in this paper only the black-box attack. White-box attacks are not considered in our scenario.
\section{III. Methodology}
To produce the adversarial samples using the GAN architecture, we adopt the approach which is similar to ADV-GAN \cite{xiao2018generating}. Based on the data set, we first train several machine-learning-based classifiers as the target models or black-box classifiers using the ground truth labels from the data set. The classifiers can be deep models or other non-deep models like random forest, decision tree and so on. Then a GAN-based architecture is build, in which the generator's input is the malicious samples or the samples that are to disguise, and the discriminator is fed with the malicious samples added with from the generator produced perturbations, and the task of the discriminator is to distinguish it whether the samples are malicious or not or the target type or not.
\begin{figure}[h]
    \centering
    \includegraphics[width=85mm]{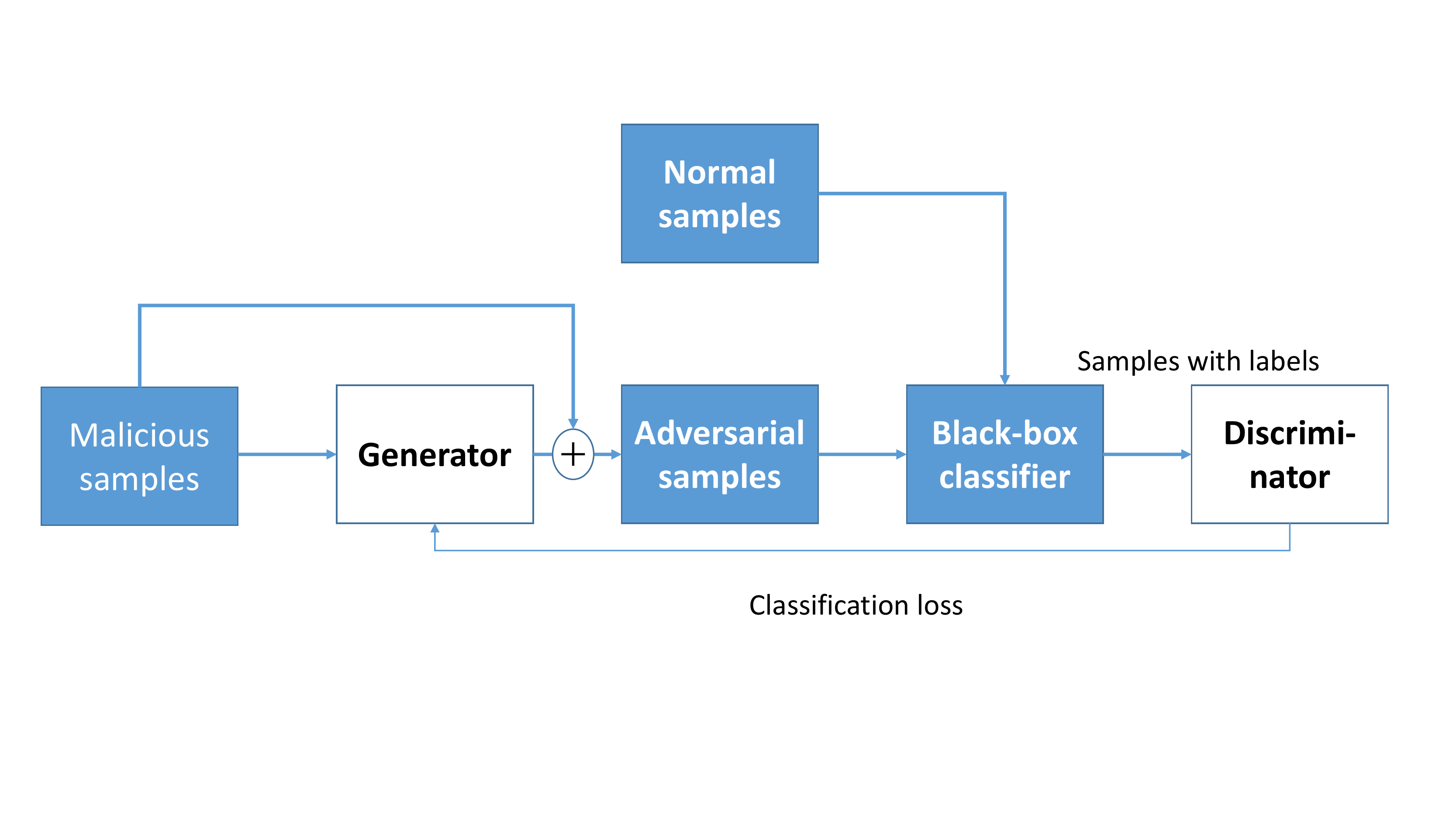}
    \label{GAN Framework}
    \caption{GAN Framework}
\end{figure}

The noise can also be the part of the input of the generator to exploit better results. In our scenario, we do not use the noise in the input of the generator because the results can be more stable and reproduceable without adding the noise. The output of the generator network is the perturbation, whose feature dimensions can be different with the feature dimensions of the sample. If so, padding zeros of the output of the generator to the same dimension as the original sample is needed. The produced perturbations are added on the original samples and the elements of the adversarial samples should be limited in the valid range and then fed into the black-box classifier with the normal samples. The predictions of the black-box classifier for the samples are regarded as the labels for the discriminator, which can be trained according to these labels to get closer and closer to the black-box classifier. The above mentioned procedure is done recursively many times through the GAN architecture until this min-max game finally comes to a equilibrium, which means producing the perturbation that can disguise the original sample and fool the black-box classifier.
\begin{equation}
\centering
\begin{aligned}
\mathop{min}\limits_{G} \mathop{max}\limits_{D}V(D,G)=E_{x\sim p_{data}(x)}[logD(x)]\\+E_{z\sim p_{z}(z)}[log(1-D(G(z)))]
\end{aligned}
\label{GAN formular}
\end{equation}
\begin{figure}[h]
    \centering
    \includegraphics[width=75mm]{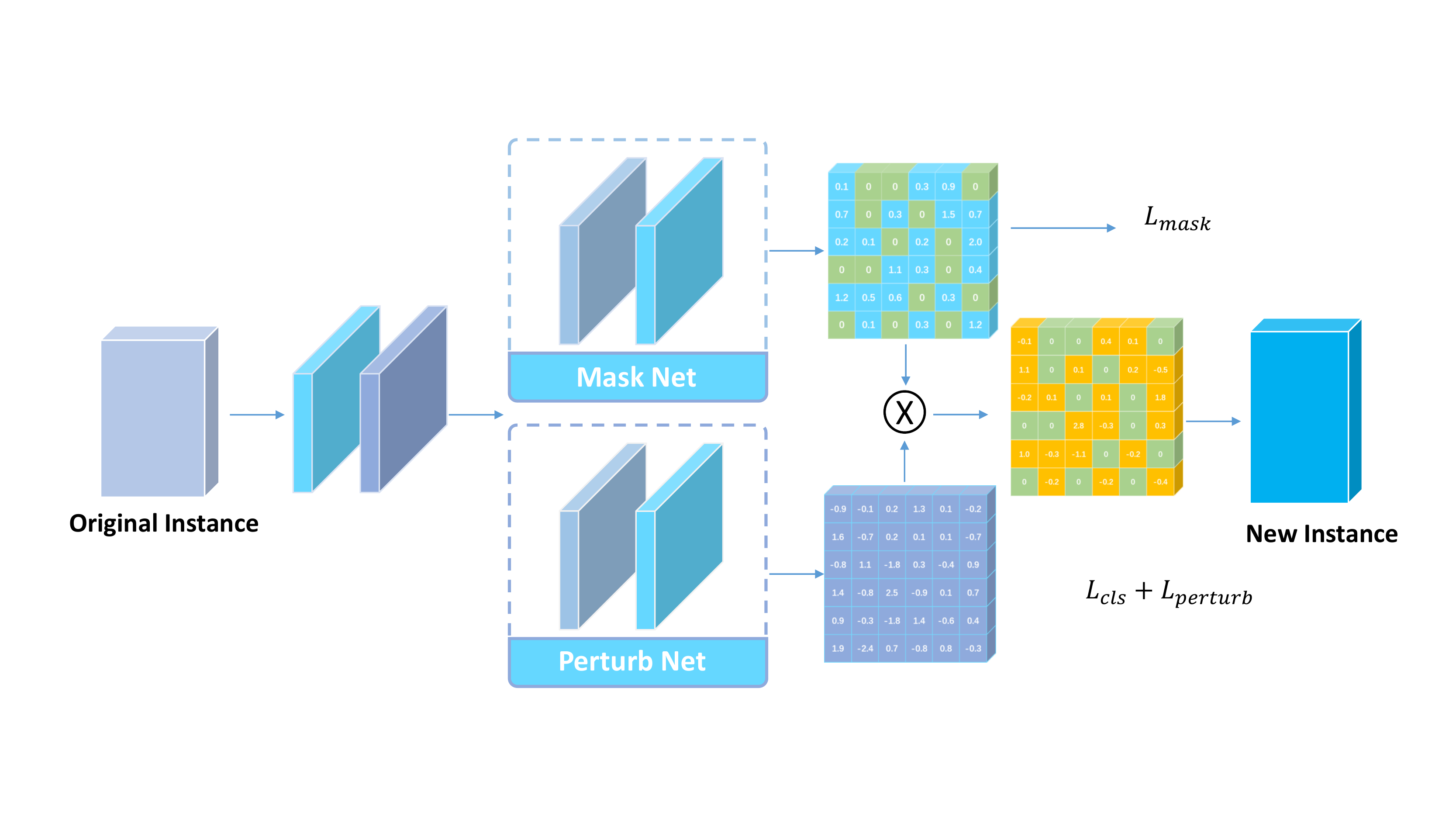}
    \label{Generator Network}
    \caption{Generator Network}
\end{figure}

Based on the basic GAN Formular \ref{GAN formular}, there are also extra constraints for the perturbation, the amplitude and the length of the perturbations, which are $length(p(X))<\epsilon_{1}$, $abs(p(X))<\epsilon_{2}$ as above mentioned. To address this constraint problem, we design a new neural network for the generator which is composed of two parallel parts: the mask network and the perturbation network. The input of generator is a feature vector, which is put into several neural layers to get latent variable $z$ of the original sample, and then the latent variable $z$ is fed into the mask network and the perturbation network respectively, and then through the dot production of two vectors of outputs of two parallel networks, we get the perturbation of the original sample. The sample is then added on the original sample to get the final attack sample.

The task of the mask network is to select the features of the perturbation that matter to the classifier. Unlike the traditional binary mask mechanism, we confine the output of mask network in the range $[0, +\infty)$ using ReLU non-linear module instead of the binary feature as in this way the gradient can be backpropagated through the mask network to the front neural layers to train the mask network. There is no labels for the mask network like in computer vision \cite{he2017mask}, therefore we train the mask network through the predefined the mask sparsity losses. The unimportant features of perturbations can be presented by zeros in the output. The output of perturb network is multiplied with the output of the mask network to produce the perturbation. And the mask network's parameters are updated by the GAN structure.

The generator losses are defined as follows:
\begin{equation}
\centering
\begin{aligned}
 L_{generator} = L_{clf} + L_{perturb} + L_{mask}\\
\end{aligned}
\end{equation}
where different losses demonstrate different meanings:
\begin{equation}
\centering
\begin{aligned}
L_{clf} = cel(f(X),f(X'))
\end{aligned}
\end{equation}
\begin{equation}
\centering
\begin{aligned}
L_{perturb} = mean(abs(p(X))) = L_1
\end{aligned}
\end{equation}
\begin{equation}
\centering
\begin{aligned}
L_{mask} = sum(abs(sign(p(X)))) = L_0
\end{aligned}
\end{equation}
Here $L_{clf}$ is the cross entropy loss of classification, $L_{perturb}$ is the loss of the amplitude of perturbations, the $L_1$ regularizer, $L_{mask}$ represents the loss of no-zero-features' length of perturbations the $L_0$ regularizer.
\begin{figure}[htbp]
  \centering
  \includegraphics[width=80mm]{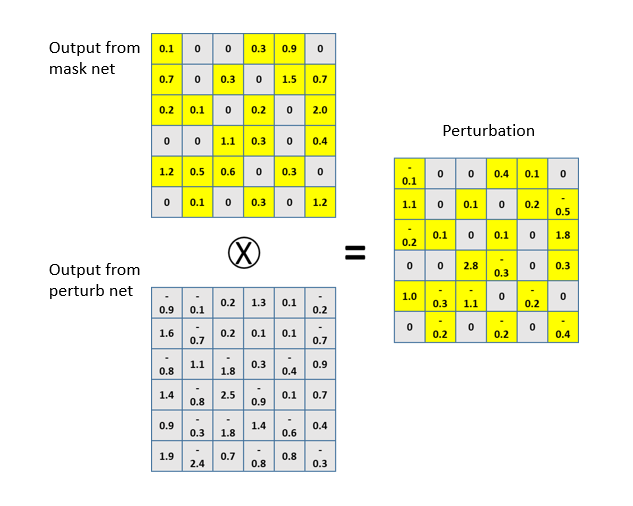}
  \caption{Dot production of two outputs of two nets}
  \label{fig_sim}
\end{figure}

Because of the two parallel networks in the generator network, it is hard to train FFA-GAN networks. This parallel structure brings the unstabilization during the training. So FFA-GAN needs a specific training method. First of all, the loss of $L_{mask}$ should be set a small value, so that the GAN can easily find the perturbation in a very wide range. Then, we increase the weight of $L_{mask}$ gradually to reduce the length of the changed feature number. Finally, we set a stop training condition, which is over a threshold bypass rate and meanwhile maintains the changed features of all features smaller than $\epsilon$, a predefined small value. The principle is that we first explore a good result in a relatively wide range, which means the great number of the changed features is tolerant in the early training phase. Then the range of the changed features is optimized as small as possible to exploit a better result. These two phases can be regarded as "the exploration phase" and "the exploitation phase" respectively.
\section{IV. Experiment}
To test the performance of the proposed FFA-GAN approach, four experiments are made on the structured and unstructured data sets.
\subsection{Structured Data Set}
\subsubsection{i. KDD-Cup 1999}
\noindent KDD-Cup 1999 \cite{stolfo2000cost} is a data set used for a big data analysis competition on the Fifth International Conference on Knowledge Discovery and Data Mining in 1999. The data were collected by MIT Lincoln Labs for the 1998 DARPA Intrusion Detection Evaluation Program. The Intrusion Detection System is a tool in the network system, which can surveille and analyse the status of the network system to detect and evaluate the possible and potential attack samples. Each data sample on KDD-Cup 1999 contains 41 features extracted from nine weeks' attacks and the normal traffic of the raw TCP dump data for a local-area network (LAN) built by DARPA, of which 9 features are symbolic and 32 features are continuous. The attack samples are composed of four categories: DOS (denial-of-service), R2L (unauthorized access from a remote machine), U2R (unauthorized access to local superuser privileges) and Probing (surveillance and other probing).
\begin{figure}[h]
  \centering
  \includegraphics[width=90mm]{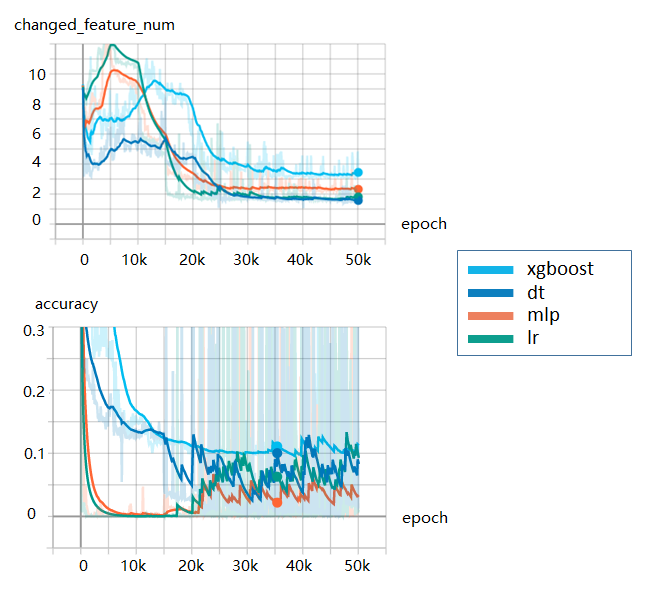}
  \caption{Training curves of FFA-GAN for the classifiers on KDD-Cup 1999}
  \label{fig_sim}
\end{figure}

The experiments of FFA-GAN on the data set KDD-Cup 1999 are made as follows: First of all, we preprocess the raw data by scaling the features in the range [0,1]. And then we randomly split the data set into two parts: 4/5 and 1/5 of all data as the training data and the testing data respectively. Based on these, four classifiers are trained to classify the data using different algorithms as the target black-box models, which are Decision Tree(\cite{safavian1991survey}), Logic Regression(\cite{ruczinski2003logic}), Multi-Layer Perception(\cite{longstaff1987pattern}) and XGBOOST(\cite{chen2016xgboost}).

The 32 continuous features are chosen for the output of the perturbation of the generator in the FFA-GAN, which are more changeable and meaningful than the other 9 symbolic features in the reality. As the dimensions of the output of the generator are smaller than the dimensions of the original sample, padding zeros is needed. Figure \ref {fig_sim} shows the training curves for the xgboost classifier. We set different weights in the epoch of 5000, 7500, 10000, 15000 and 20000 to train the FFA-GAN. The targeted attack for the normal label makes sense in this scenario because in reality, the attack samples are what we want to disguise as the normal samples to fool the classifier.

It is obvious that in Figure \ref{fig_sim} the number of changed features decreases, meanwhile the accuracy of the four classifiers falls from over 99\% to less than 10\%. The first graph shows very clear that in each predefined epoch (5000, 7500, 10000, 15000 and 20000), there are a significant change of the number of changed features. The score curves gradually decrease and stay stabile no matter the weights change, which demonstrates the stabile results of the training procedure at the end.
\begin{table}[h]
\small
\caption{FFA-GAN and IDS-GAN targeted attack results for different classifiers on KDD-Cup 1999}
\begin{center}
\begin{tabular}{|p{1.5cm}<{\centering}|p{0.7cm}<{\centering}|p{0.7cm}<{\centering}|p{0.7cm}<{\centering}|p{0.7cm}<{\centering}|p{0.7cm}<{\centering}|p{0.7cm}<{\centering}|}
\hline
\multirow{2}{*}{Classifier} & \multirow{2}{*}{acc} & \multicolumn{2}{c|}{ FFA-GAN } & \multicolumn{2}{c|}{IDS-GAN }\\
\cline{3-6} &  & acc* & len  & acc* & len \\ \hline
dt & 99.9\%  &8.1\%& \textbf{1.6}  &0.4\%& \multirow{4}{*}{\textbf{20.5}} \\
\cline{1-5}lr & 94.2\%  &6.0\%& \textbf{1.7}  &0.6\%& \\
\cline{1-5}mlp & 99.9\%  &3.0\%& \textbf{2.2}  &0.7\%& \\
\cline{1-5}xgboost & 99.6\%  &10.3\%& \textbf{3.2}  &-\%&  \\ \hline
\end{tabular}
\end{center}
\label{table1}
\footnotesize{("acc*" means the classifier's accuracy after the adversarial attack. "len" means the length of non-zero features of the perturbation. The attack IDS-GAN contains no "Probing".)}
\end{table}

The total results of the FFA-GAN on KDD-Cup 1999 for the four classifiers can be concluded as shown in the Table \ref{table1}. From the table, we can see a enormous accuracy rate fall after using FFA-GAN and only very few features need to be changed. For the approach IDS-GAN \cite{lin2018idsgan}, the average changed number is over 20 features, ca. 45\% of all features. In contrast, we use only 2-3 features (less than 7.5\% of all features) to achieve a high bypass rate as in IDS-GAN.
\subsubsection{ii. CIC-IDS 2017}
Like KDD-Cup 1999, CIC-IDS 2017 \cite{sharafaldin2018toward} is also a data set about the intrusion detection system. The difference lies in that CIC-IDS 2017 is a relatively new data set and contains totally 78 features and 8 attack types for each sample. Excluding the discrete features, we implement FFA-GAN only on the continuous features. From the Table \ref{table2}, we can see the better results on the CIC-IDS 2017 than on KDD-Cup 1999. We can also see a enormous accuracy fallen after using FFA-GAN and only very few features need to be changed and meanwhile FFA-GAN achieves a very high bypass rate. This is also a proof that FFA-GAN generalizes on different data set with the low dimensional structured data.
\begin{table}[htbp]
\small
\caption{FFA-GAN targeted attack results for different classifiers on CIC-IDS 2017}
\begin{center}
\begin{tabular}{|p{1.5cm}<{\centering}|p{1.5cm}<{\centering}|p{1.5cm}<{\centering}|p{1.5cm}<{\centering}|}
\hline
Classifier & acc & acc*& len\\ \hline
dt & 99.0\%  &4.0\%& \textbf{3.2}\\ \hline
lr & 91.3\%  &0.1\%& \textbf{2.1}\\ \hline
mlp & 99.1\%  &0.5\%& \textbf{2.3}\\ \hline
xgboost & 99.6\%  &0.1\%& \textbf{2.2}\\ \hline
\end{tabular}
\end{center}
\label{table2}
\footnotesize{("acc*" means the classifier's accuracy after the FFA-GAN attack. "len" means the length of non-zero features of the perturbation.)}
\end{table}
\subsection{Unstructured Data Set}
\subsubsection{i. MNIST}
\begin{figure}[h]
  \centering
  \includegraphics[height=125mm]{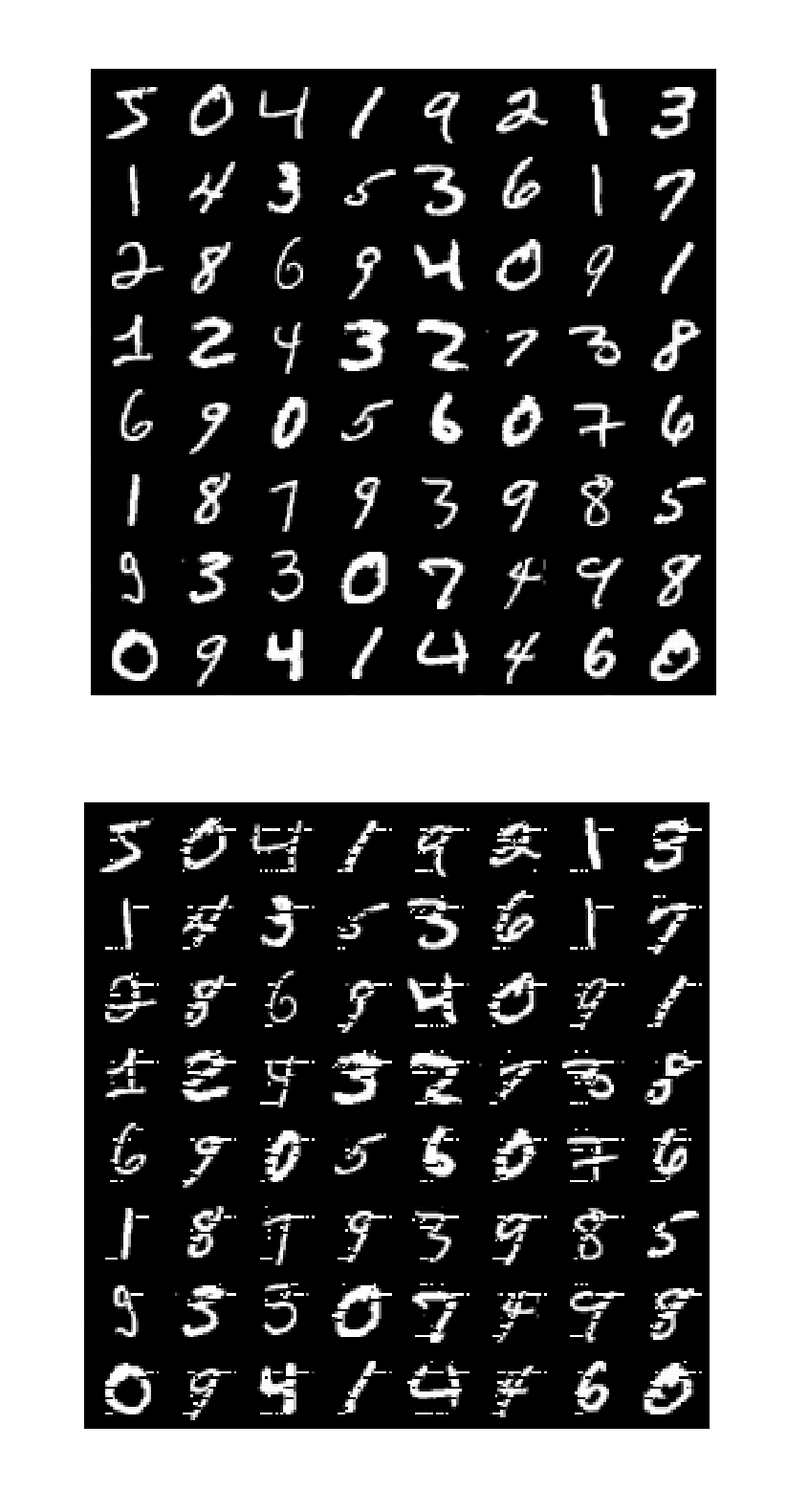}
  \caption{Attack Examples in MNIST}
  \footnotesize{(The upper images are the original samples. The bottom images are the FFA-GAN adversarial samples which are all classified as the target arabic number 5 based on the target resnet34-classifier.)}
  \label{mnistattack}
\end{figure}

MNIST data set \cite{lecun1998gradient} is a classic data set in the domain of image processing. The images were collected by National Institute of Standards and Technology (NIST) and contains 60,000 training images and 10,000 testing 28 pixels * 28 pixels images, which are composed of 10 arabic number. The target classifiers are vgg16 \cite{simonyan2014very} and resnet34 \cite{he2016deep}. Both have the classification accuracy over 99.0\% on MNIST data set. As the untargeted attack is easier than the targeted target, we only implement FFA-GAN to attack the classifiers in the targeted way to test the FFA-GAN approach. Table \ref{table3} shows the different approaches' results of FFA-GAN based on the different target classifier. The final average perturbations' length of FFA-GAN is smaller than 3\% of all pixels and ca. 99.5\% samples can be classified incorrectly as the predefined target type. In contrast, the ADV-GAN needs to change a great lot of features to achieve the high bypass rate, for the one pixel attack approach we need to use 30 pixels attack to achieve ca. 83\% bypass rate based on the vgg16-classifier. Figure \ref{mnistattack} demonstrates the attack examples on MNIST using FFA-GAN.
\begin{table}[htbp]
\small
\caption{FFA-GAN, ADV-GAN and One Pixel Attack testing results on MNIST}
\begin{center}
\begin{tabular}{|p{1.1cm}<{\centering}|p{0.6cm}<{\centering}|p{0.6cm}<{\centering}|p{0.6cm}<{\centering}|p{0.6cm}<{\centering}|p{0.6cm}<{\centering}|p{0.6cm}<{\centering}|p{0.6cm}<{\centering}|p{0.6cm}<{\centering}|}
\hline
\multirow{2}{*}{Classifier} & \multirow{2}{*}{acc} & \multicolumn{2}{c|}{ FFA-GAN } & \multicolumn{2}{c|}{ADV-GAN } & \multicolumn{2}{c|}{OP Attack}\\
\cline{3-8} &  & acc* & len  & acc* & len & acc* & len \\ \hline
vgg16-targeted & 99.5\%  &0.3\%& \textbf{25.6} &0.1\%& \textbf{525.6} &16.0\%& \textbf{30}\\ \hline
resnet34-targeted & 99.3\%  &0.5\%& \textbf{24.4} &0.2\%& \textbf{530.1} &-\%& \textbf{-}\\ \hline
\end{tabular}
\end{center}
\label{table3}
\footnotesize{("acc*" means the classifier's accuracy after the adversarial attack. "len" means the length of non-zero features of the perturbation.)}
\end{table}
\subsubsection{ii. CIFAR-10}
\begin{figure}[h]
  \centering
  \includegraphics[height=75mm]{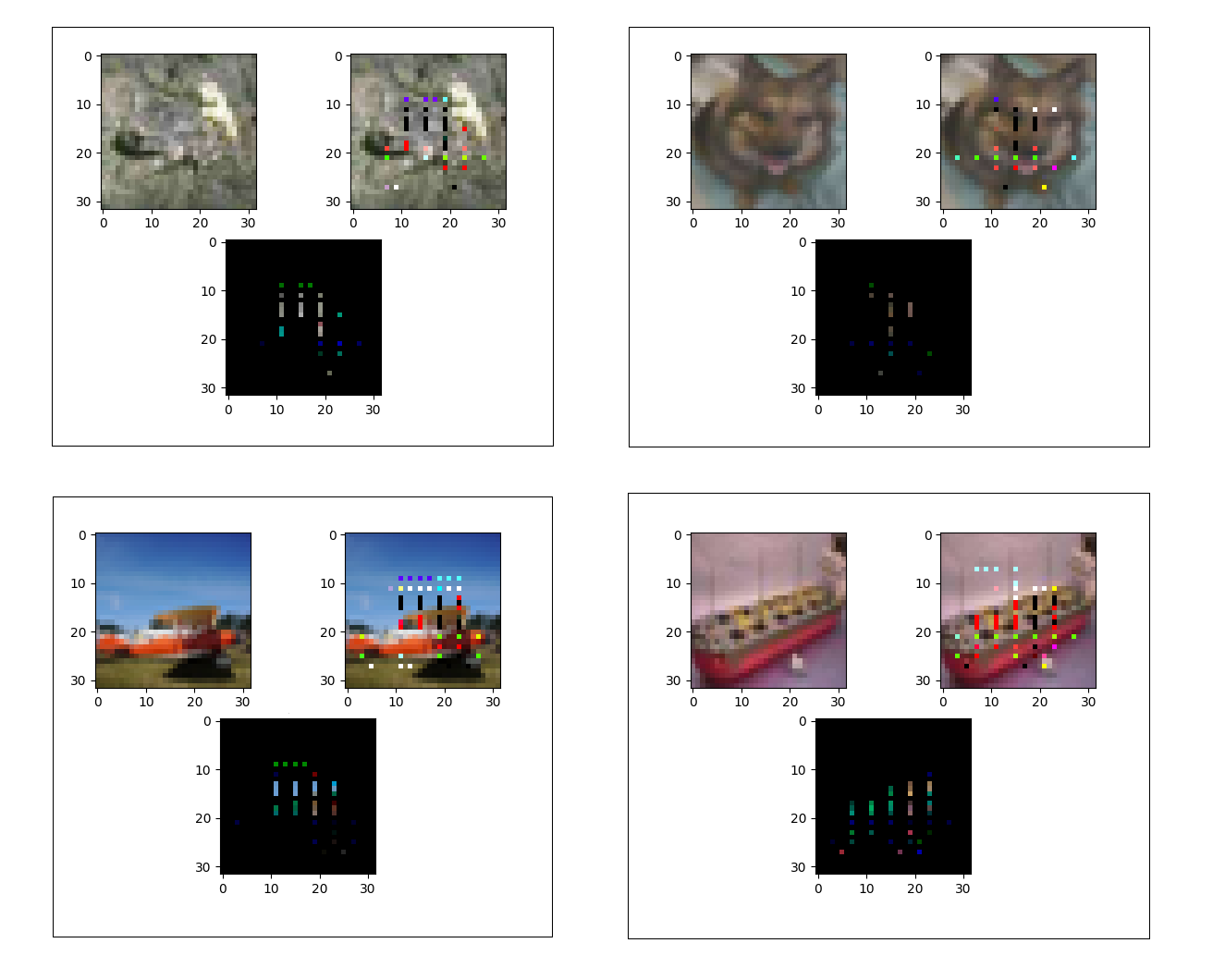}
  \caption{Attack Examples in CIFAR-10}
  \footnotesize{(In every frame, the left image is the original sample, the right image is the FFA-GAN adversarial sample, the bottom image is the perturbation. The above examples are all classified as the target type TRUCK based on the target vgg16-classifier.)}
  \label{cifarattack}
\end{figure}

CIFAR-10 data set \cite{krizhevsky2009learning} is an image data set, which were collected by Alex Krizhevsky, Vinod Nair, and Geoffrey Hinton. There are totally 60000 images in 10 classes, with 6000 32 pixels x 32 pixels colour images per class. The data set is composed of five training batches and one test batch, each with 10000 images. The difference between the CIFAR-10 data set and the MNIST data set is their channels: the image of CIFAR-10 has 3 channels and the image of MNIST has only one channel, which means it is more difficult to train FFA-GAN on the data set CIFAR-10 than on MNIST. Like the other experiments, the classifiers for the data set are firstly trained. Here we also use vgg16 and resnet34, which both have over 93\% classification accuracy on CIFAR-10. After using the FFA-GAN approach, the accuracy rate of vgg16 and resnet34 has fallen to the 40\% and 30\% respectively, meanwhile only ca. 4.5\% of all pixels are perturbed. ADV-GAN \cite{xiao2018generating} uses over 50\% of all pixels to achieve a high bypass rate. One pixel attack approach \cite{su2019one} needs to use 30-pixels attack to reduce the accuracy of the sample to 17.1\% averagely. Figure \ref{cifarattack} are the successful attack examples of FFA-GAN.
\begin{table}[htbp]
\small
\caption{FFA-GAN, ADV-GAN and One Pixel Attack testing results on CIFAR-10}
\begin{center}
\begin{tabular}{|p{1.1cm}<{\centering}|p{0.6cm}<{\centering}|p{0.6cm}<{\centering}|p{0.6cm}<{\centering}|p{0.6cm}<{\centering}|p{0.6cm}<{\centering}|p{0.6cm}<{\centering}|p{0.6cm}<{\centering}|p{0.6cm}<{\centering}|}
\hline
\multirow{2}{*}{Classifier} & \multirow{2}{*}{acc} & \multicolumn{2}{c|}{ FFA-GAN } & \multicolumn{2}{c|}{ADV-GAN } & \multicolumn{2}{c|}{OP Attack}\\
\cline{3-8} &  & acc* & len  & acc* & len & acc* & len \\ \hline
vgg16-targeted & 93.2\%  &53.1\%& \textbf{48.0} &14.3\%& \textbf{551.3} &17.1\%& \textbf{30}\\ \hline
resnet34-targeted & 94.2\%  &60.2\%& \textbf{51.4} &13.2\%& \textbf{547.7} &-\%& -\\ \hline
\end{tabular}
\end{center}
\footnotesize{("acc*" means the classifier's accuracy after the adversarial attack. "len" means the length of non-zero features of the perturbation.)}
\label{table4}
\end{table}
\subsection{Summary}

\begin{table}[h]
\small
\caption{One Pixel Attack and FFA-GAN results for testing 100 images on MNIST}
\begin{center}
\begin{tabular}{|p{1.2cm}<{\centering}|p{0.5cm}<{\centering}|p{0.5cm}<{\centering}|p{0.5cm}<{\centering}|p{0.5cm}<{\centering}|p{0.5cm}<{\centering}|p{0.5cm}<{\centering}|p{0.4cm}<{\centering}|p{0.4cm}<{\centering}|}
\hline
\multicolumn{7}{|c|}{ OP Attack }& \multicolumn{2}{c|}{FFA-GAN} \\ \hline
iteration & acc-1 & acc-5 & acc-10 & acc-20 & acc-30 & T(s) & acc & T(s)\\ \hline
75-vgg16 & 100\%  &91\%& 88\% &64\%& 55\% &466& \multirow{3}{*}{0.1\%}& \multirow{3}{*}{0.1}\\
\cline{1-7}150-vgg16 & 99\%  &84\%& 77\% &55\%& 43\% &896& & \\
\cline{1-7}300-vgg16 & 97\%  &84\%& 60\% &39\%& 30\% &2532& & \\ \hline
75-resnet34 & 96\%  &48\%& 36\% &22\%& 15\% &305& \multirow{3}{*}{0.2\%}& \multirow{3}{*}{0.1}\\
\cline{1-7}150-resnet34 & 95\%  &24\%& 18\% &11\%& 5\% &615& & \\
\cline{1-7}300-resnet34 & 98\%  &16\%& 4\% &2\%& 2\% &1078& & \\ \hline
\end{tabular}
\end{center}
\footnotesize{("x-model" for OP Attack means the execution optimization iteration time based on the target model. "acc-x" for OP Attack means the classifier's accuracy after the x-pixels attack. Time is calculated per one image. The experiments are made on the computer with Win10, Memory 64GB, GPU 2080 and CPU i7-8700k.)}
\label{table5}
\end{table}
From the four experiments, we can draw the conclusion that FFA-GAN is an effective and robust approach to generate the adversarial samples on the structural and unstructural data set. In comparison with the GAN-based approaches IDS-GAN \cite{lin2018idsgan} and ADV-GAN \cite{xiao2018generating}, FFA-GAN can obviously reduce the length of the needed perturbations and meanwhile keep a very high bypass rate as shown in Table \ref{table1}, \ref{table2}, \ref{table3}, \ref{table4}. In comparison with the traditional non-learning approaches One-pixel-attack \cite{su2019one}, FFA-GAN can generate the adversarial samples using only once forward computing for the FFA-GAN after training. However, the other non-learning approaches need to use the classifier many times until the successful adversarial samples are found. For instance, one pixel attack approach spends much more time using the target classifier per samples to get the final results on MNIST as shown in Table \ref{table5}. Besides, the bypass rate FFA-GAN is better than One-pixel-attack. In the term of the sparsity, FFA-GAN's result is relatively worse than one-pixel attack in CIFAR-10. The reason might be that the mask, which decides the sparsity of the perturbation, is trained by the neural network and it is hard to train the $L_0$ regularizer as it is an NP-problem. Therefore, the hyperparameters of the losses' weights, which are manually predefined, play an important role in FFA-GAN.

\section{V. Conclusion and Future Work}
In this paper, we proposed a Few-Feature-Attack GAN approach to generate adversarial examples to fool machine-learning models. FFA-GAN has a significant time-consuming advantage than the non-learning adversarial samples approaches and a better non-zero-features performance than the GAN-based adversarial sample approaches. Through the introduction of the mask mechanism, the number of changed features of the perturbation is constraint as few as possible. And we define three losses of the generator network based on the GAN architecture and change the weights of the losses during the training to find a better result in a wide range firstly and minimize the range later. Experiments on the structured data sets and unstructured data sets results show the good performance of the FFA-GAN approach. For the future work, the change of the weights of the losses during training can be optimised using the optimization algorithms such as Population based Training method \cite{jaderberg2017population}. PBT can parallelly train GANs as a population in large scale and optimise the hyperparamers of the different losses through evolution algorithms. Besides, the work on exploiting the performance of FFA-GAN in the image with high resolutions is also worth to be studied in the future.

\section*{VI. Acknowledgment}
 We would like to extend the deep gratitude to all those who have supported us practically, cordially and selflessly in writing this paper.



\bibliographystyle{IEEEtran}
\bibliography{Bibliography-File}
\end{document}